\documentclass[letterpaper, 10 pt, conference]{ieeeconf}  

\IEEEoverridecommandlockouts                              

\usepackage{graphics} 
\usepackage{epsfig} 
\usepackage{xcolor}
\usepackage[table]{xcolor}
\definecolor{grey}{rgb}{0.5,0.5,0.5}
\usepackage{booktabs}
\usepackage{subcaption}
\usepackage{hyperref}
\usepackage{amsmath}
\usepackage{algorithm}
\usepackage{algpseudocode}
\usepackage{float}
\usepackage{url}
\usepackage{amsmath} 
\usepackage{amssymb}  
\usepackage[
    style=ieee,
    natbib=true,
    citestyle=numeric-comp,
    doi=false,
    isbn=false,
    url=false]{biblatex} 

\newcommand{\eg}{e.\,g.\,}

\newcommand\etc{etc\@ifnextchar.{}{.\@}}

\definecolor{deepred}{RGB}{196, 49, 25}
\definecolor{tablepeachdark}{RGB}{255,215,196}
\definecolor{tablepeach}{RGB}{255, 240, 235}

\definecolor{tablepurpledark}{RGB}{224,190,240}
\definecolor{tablepurple}{RGB}{248,235,252}

\definecolor{tableblue}{RGB}{235,241,255}
\definecolor{tablebluedark}{RGB}{197,214,255}

\title{\LARGE \bf
Behavior Foundation Model for Humanoid Robots}
\author{Weishuai Zeng$^{1,5}$, Shunlin Lu$^{2,5}$, Kangning Yin$^{3,5}$, Xiaojie Niu$^{5}$, Minyue Dai$^{4,5}$, Jingbo Wang$^{5}$, Jiangmiao Pang$^{5}$
\\
$^{1}$Peking University, $^{2}$The Chinese University of Hong Kong, Shenzhen \\$^{3}$Shanghai Jiaotong University, $^{4}$Fudan University, $^{5}$Shanghai Artificial Intelligence Laboratory}

\begin{document}


\makeatletter
\renewcommand{\@maketitle}{%
  \newpage
  \null
  \begin{center}%
    {\LARGE \bf \@title \par}%
    \vskip 0.5cm %
    \@author
  \end{center}
  \vspace{-0.1em} 
  \begin{center}
    \includegraphics[width=1.0\linewidth]{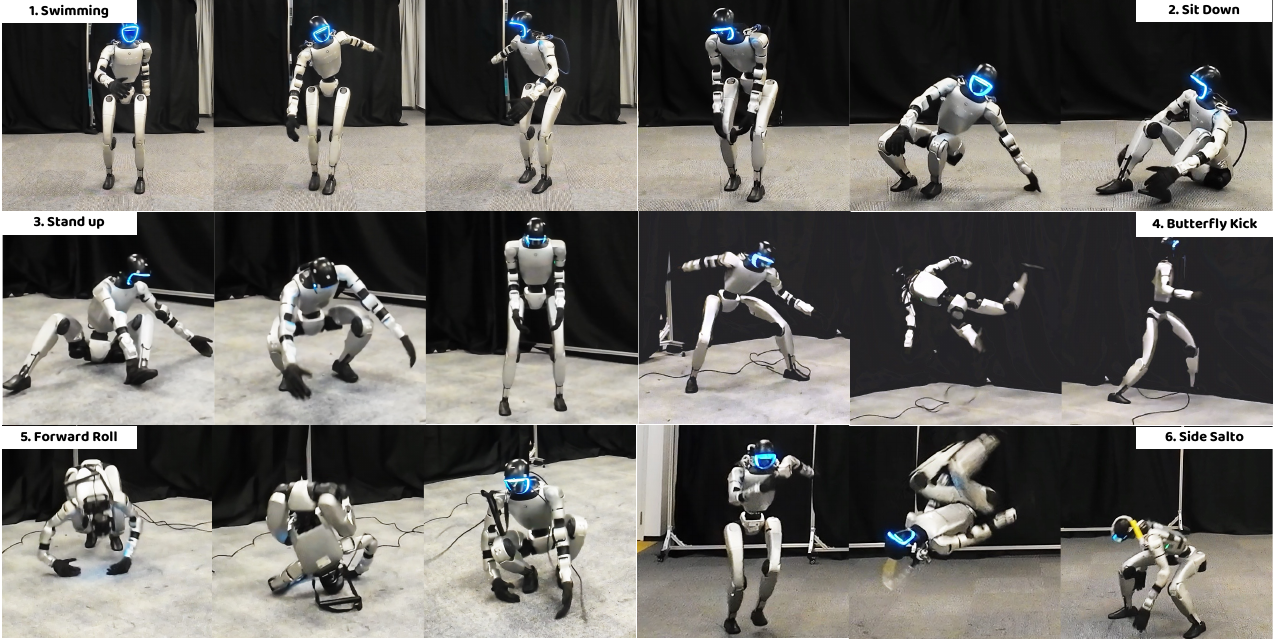}
    \captionof{figure}{
      \textbf{Behavior Foundation Model} enables humanoid robots to perform a variety of behaviors in a zero-shot manner, including (1) a swimming pose, (2) sitting down on the ground, (3) standing up from the ground, and (4) butterfly kick. It also facilitates efficient acquisition of new behaviors such as (5) a forward roll and (6) a side salto.}
    \label{fig:teaser}
  \end{center}
  \setcounter{figure}{1}
  \vspace{-2mm}
}
\makeatother

\maketitle

\thispagestyle{empty}
\pagestyle{empty}

\begin{abstract}
Whole-body control (WBC) of humanoid robots has witnessed remarkable progress in skill versatility, enabling a wide range of applications such as locomotion, teleoperation, and motion tracking. Despite these achievements, existing WBC frameworks remain largely task-specific, relying heavily on labor-intensive reward engineering and demonstrating limited generalization across tasks and skills. These limitations hinder their response to arbitrary control modes and restrict their deployment in complex, real-world scenarios. To address these challenges, we revisit existing WBC systems and identify a shared objective across diverse tasks: the generation of appropriate behaviors that guide the robot toward desired goal states. Building on this insight, we propose the {Behavior Foundation Model (BFM)}, a generative model pretrained on large-scale behavioral datasets to capture broad, reusable behavioral knowledge for humanoid robots. BFM integrates a masked online distillation framework with a Conditional Variational Autoencoder (CVAE) to model behavioral distributions, thereby enabling flexible operation across arbitrary control modes and efficient acquisition of novel behaviors without retraining from scratch. Extensive experiments in both simulation and on a physical humanoid platform demonstrate that BFM generalizes robustly across diverse WBC tasks while rapidly adapting to new behaviors. These results establish BFM as a promising step toward a foundation model for general-purpose humanoid control. Videos and supplementary materials are available at: \href{https://bfm4humanoid.github.io}{\textcolor{deepred}{\texttt{bfm4humanoid.github.io}}}.
.

\end{abstract}

\section{INTRODUCTION}
\vspace{-1.0mm}

Humanoid robots are capable of executing a wide range of whole-body control (WBC) tasks, including language interaction~\cite{jiang2024harmon,shao2025langwbc}, human teleoperation~\cite{he2024omnih2o, ben2025homie, lu2025mobile} and whole-body motion tracking~\cite{he2025asap,chen2025gmt}. Despite this versatility, most existing WBC systems are typically designed for specific control modes. For example, a motion tracking policy accepts only reference motions, whereas a locomotion policy responds exclusively to velocity commands. Such rigid specialization of control modes hinders cross-task generalization: a locomotion policy, for instance, cannot directly exploit reference motions for whole-body tracking. Recent studies have attempted to support multiple control modes with mask strategies. However, these approaches are either confined to simplified virtual avatars~\cite{tessler2024maskedmimic} or prioritize a fixed set of control modes~\cite{he2025hover}, thus failing to accommodate arbitrary control modes on real humanoid robots. We argue that the root cause of this limitation lies in the absence of a unified formulation across diverse tasks. To address this challenge, we revisit the design of existing WBC systems and make a key observation: although control modes differ, the resulting outcomes of these systems, whether walking or dancing, are all fundamentally \textbf{behaviors} of humanoid robots, which naturally serve as a unified formulation across diverse tasks. Under this perspective, task-specific control modes, whether velocity commands, VR signals or reference motions can all be interpreted as distinct specifications of the behaviors.

One behavior can often be specified through multiple control modes. A simple yet illustrative example is that the behavior of walking forward can be learned under both locomotion and motion tracking settings. This observation suggests that, despite variations in control modes, existing WBC systems ultimately pursue a shared objective: the generation of appropriate behaviors. Such a perspective motivates the decoupling of behaviors from control modes, thereby shifting the paradigm from isolated task learning toward holistic behavior learning. Inspired by the success of foundation models in other domains~\cite{radford2019language}, if we may pretrain a foundation model on large and diverse behavioral datasets, it may encode a broad spectrum of behavioral knowledge, which can then be applied to a variety of downstream tasks. 

To this end, we introduce the \textbf{Behavior Foundation Model (BFM)} for humanoid robots, a generative model pretrained on large-scale behavioral datasets to capture broad and reusable behavioral knowledge. BFM can be directly steered by diverse control modes to accomplish corresponding tasks, while also enabling the efficient acquisition of novel behaviors without the need of retraining from scratch. In general, BFM establishes a flexible and generalizable framework for humanoid control, highlighting its potential as a foundation for the next generation of WBC system design.

To realize this vision, we first adopt motion imitation as a common abstraction of behaviors to train a proxy agent in simulation. Then, we pretrain BFM using a masked online distillation framework combined with a Conditional Variational Autoencoder (CVAE)~\cite{higgins2017beta}, which provides a versatile control interface capable of supporting diverse control modes as well as a structured latent space that facilitates both behavior composition and modulation. Furthermore, we integrate residual learning~\cite{he2016deep} into our framework, enabling efficient acquisition of novel behaviors by leveraging the behavioral knowledge already encoded in the BFM. As presented in Figure \ref{fig:teaser}, the overall framework demonstrates both versatility and robustness in real-world deployment.
 
In summary, our contributions are threefold: 1) we present Behavior Foundation Model for humanoid robots, shifting the focus of humanoid control from holistic task learning to unified behavior learning; 2) we demonstrate our framework integrating masked online distillation and CVAE can be directly steered for diverse WBC tasks and may efficiently acquire new behaviors via residual learning without retraining from scratch; 3) extensive experiments in both simulation and on a real humanoid robot validate expressiveness and effectiveness of our BFM, highlighting its potential as a foundation for developing general-purpose humanoid robots.

\section{Related Work}

\subsection{Humanoid Whole-body Control}

Existing WBC systems for humanoids can be generally categorized by their control modes ranging from abstract to concrete. The most abstract control mode is natural language~\cite{jiang2024harmon, shao2025langwbc} which may correspond to a series of feasible behaviors that all satisfy the given instructions. A more concrete and widely adopted control mode, especially for locomotion and loco-manipulation, involves the velocity commands and base height commonly combined with other signals like gait and posture~\cite{xue2025unified}
, upper-body joint positions from exo-skeleton~\cite{ben2025homie} or VR devices with inverse kinematics (IK)~\cite{lu2025mobile}. Besides processing VR signals with IK, other teleoperation systems~\cite{he2024omnih2o} directly 
map kinematic data from VR controllers to the humanoid, enabling highly expressive and accurate whole-body control. The most concrete control mode for existing WBC systems is motion tracking which provides nearly complete information about the reference pose either from offline datasets~\cite{chen2025gmt} or online motion capture systems~\cite{ze2025twist}. While these systems produce impressive results, their control mode is often determined at design time and therefore lack cross-task generalization. HOVER~\cite{he2025hover} attempts to address this by employing a unified policy with a masking strategy to achieve multi-modal control, demonstrating versatile humanoid control across diverse WBC tasks.

\subsection{Behavior Foundation Model}
Recent advances in reinforcement learning has led to Behavior Foundation Models, which exhibit versatile behavior generation and strong generalization across diverse tasks. Existing works have implemented BFM from distinct perspectives. Motivo~\cite{tirinzoni2025zero} and its series of works use forward-backward representations to enable unsupervised learning on reward-free transitions, yielding near-optimal policies for zero-shot inference across diverse tasks. MaskedMimic~\cite{tessler2024maskedmimic} and HOVER~\cite{he2025hover} employ online masked distillation to train goal-conditioned policies, which also allows zero-shot generalization across tasks and contexts. Our framework draws inspiration from the latter line of work but differs in key aspects. HOVER's two-stage mask strategy still prioritizes specific control modes, while our BFM supports arbitrary modes through direct application of sparsity mask. Maskedmimic focuses on simplified virtual avatars and lacks the latent space analysis to clarify advantages of CVAE over other generative models, while our BFM targets real-world humanoids and reveals the latent space properties for applications like behavior composition and modulation. Through both theoretical and empirical contributions, our BFM offers a systematical analysis of the underlying unified formulation existing models have actually learned and enables a wide range of downstream applications on real humanoid robots.

\section{Proxy Agent Training}

\begin{figure*}[tbp]
    \centering
    \includegraphics[width=1.0\textwidth]{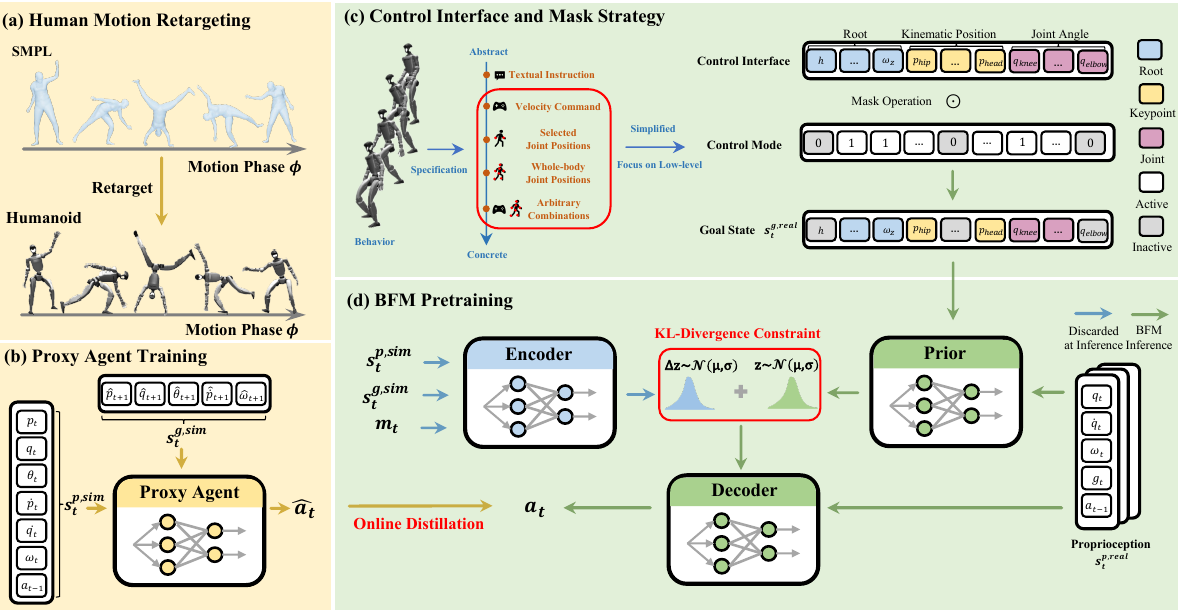}
    \caption{\textbf{Overview of BFM Implementation.} (a) Human motion dataset is retargeted to humanoid robots for proxy agent training. (b) The proxy agent is trained via motion imitation which has access to all information in simulators. (c) One behavior can be specified in multiple control modes from abstract textual instructions to concrete whole-body joint positions, resulting distinct goal states. We simplify the activation of distinct control modes to applying mask to a unified control interface. This interface is restricted to the union of root, kinematic position, and joint angle, emphasizing our focus on low-level humanoid control. (d) We model the BFM with a CVAE and employ the DAgger framework for BFM pretraining, providing a structured latent space to encode extensive behavioral knowledge.}
    \label{fig:firstpage}
    \vspace{-15pt}
\end{figure*}

\subsection{Behaviors under Reinforcement Learning Formulation}

We formulate the problem of humanoid control as a goal-conditioned reinforcement learning (RL) task, where a policy $\pi$ is trained to achieve certain objectives. The state $s_t$ comprises both the humanoid's proprioception $s_t^p$ and the goal state $s_t^g$. Using the humanoid's proprioception $s_t^p$ and the goal state $s_t^g$, the reward function is defined as $r_t=R(s_t^p, s_t^g)$ for policy optimization. The action $a_t$ represents the target joint positions for humanoids, which are then fed into the PD controller to actuate the robot's degrees of freedom. Proximal Policy Optimization (PPO) algorithm~\cite{schulman2017proximal} is employed to maximize the cumulative reward $E[\sum_{t=1}^T\gamma^{t-1}r_t]$.

With the formulation above, we define behaviors as \textit{trajectories over the humanoid’s proprioceptive states and actions}. We particularly exclude task-related goal states $s_t^g$ from trajectories, as we interpret them as external motives that drive the generation of appropriate behaviors. Such exclusion avoids pre-determination of control modes, allowing behaviors to be a unified formulation for humanoid control.

In order to distinguish the observable states during real-world deployment from the privileged states in simulators, we use $s_t^{p,sim}$ and $s_t^{g,sim}$ to represent privileged states in simulators and $s_t^{p,real}$ and $s_t^{g,real}$ to represent observable states available during real-world deployment. With such notations, the behavior is defined as trajectories over $s_t^{p,real}$ and $a_t$, expressed as $\tau = [s_1^{p,real},a_1, s_2^{p,real}\dots, s_{T-1}^{p,real},a_{T-1},s_{T}^{p,real}]$.

\vspace{-0.5em}
\subsection{Human Motion Retargeting}

Human motion dataset plays an important role in humanoid behavioral dataset preparation for their diversity and high quality. We select the publicly available AMASS dataset~\cite{mahmood2019amass} where each motion sample is parameterized by the SMPL model~\cite{loper2023smpl}. To bridge the embodiment gap between SMPL human model and humanoid robots, we employ a two-stage retargeting approach~\cite{he2024learning}. First, we optimize the shape parameter of SMPL model for humanoid robots by minimizing distances between selected links in the rest pose. Second, we optimize the humanoid's root translation, orientation and joint positions by minimizing distances between selected links throughout the whole sequence. Additional regularization terms are added to avoid aggressive behaviors and ensure temporal smoothness. 

Unlike previous works which use a motion imitator trained with privileged information in simulators to further filter the dataset, we directly use the raw dataset for BFM pretraining.
\vspace{-0.5em}
\subsection{Proxy Agent Training via Motion Imitation}

Instead of organizing behavioral dataset as offline trajectories for BFM pretraining, we train a proxy agent denoted as $\pi^{proxy}$ using motion imitation. The obtained proxy agent may provide actions given the current proprioceptive state and goal state derived from the reference motion, which may generate quantities of behavioral data by online rolling out.

\noindent \textbf{State Space Design.} The state space of the proxy agent is comprised of the privileged proprioception and goal state in simulators. The privileged proprioception for proxy agent is defined as $s_t^{p,sim}\triangleq [p_t, q_t, \theta_t, \dot{p_t}, \dot{q_t}, \omega_t, a_{t-1}]$, which contains the humanoid rigid-body position $p_t$, joint position $q_t$, orientation $\theta_t$, linear velocity $\dot{p_t}$, joint velocity $\dot{q_t}$, angular velocity $\omega_t$  and previous action $a_{t-1}$. The privileged goal state is defined as $s_t^{g,sim}\triangleq [\hat{p}_{t+1}-p_t, \hat{q}_{t+1}-q_t, \hat{\theta}_{t+1}\ominus\theta_t, \hat{v}_{t+1}-v_t, \hat{\omega}_{t+1}-\omega_t, \hat{p}_{t+1}-p_t^{root}, \hat{\theta}_{t+1} \ominus \theta_t^{root}]$, which contains the one-frame difference between the reference pose $(\hat{p}_{t+1}, \hat{q}_{t+1}, \hat{\theta}_{t+1}, \hat{v}_{t+1}, \hat{\omega}_{t+1})$ and the current pose. $p_t^{root}$ refers to the root translation and $\theta_t^{root}$ refers to the root orientation of the current pose. All these goal states are rotated to the local coordinate of the current frame.

\noindent \textbf{Reward Design and Domain Randomization.} We formulate the reward $r_t$ as a weighted sum of three components: 1) task rewards for motion imitation, 2)regularization, and 3)penalty, as detailed in Table \ref{tab:reward}. We employ curriculum learning to the regularization and penalty terms, encouraging the policy to focus on motion imitation initially and gradually leverage penalty and regularization to shape the behaviors. We also apply domain randomization during training by randomizing dynamics and applying external perturbations. Details of domain randomization are listed in table \ref{tab:dr}.

\begin{table}[tbp]
\centering
\caption{Reward Designs for Proxy Agent}
\vspace{-3pt}
\renewcommand{\arraystretch}{1.2}
\resizebox{0.98\linewidth}{!}{%
\begin{tabular}{ c  c  c  c }
\hline
Term & Weight & Term & Weight \\ \hline
\multicolumn{4}{c}{\textbf{Task Reward}} \\ \hline
Body position & $1.0$ & Body position (selected keypoint) & $1.6$ \\
Body position (feet) & $2.1$ & Body rotation & $0.5$ \\
Body velocity & $0.5$ & Body angular velocity & $0.5$ \\
DoF position & $0.75$ & DoF velocity & $0.5$ \\ \hline
\multicolumn{4}{c}{\textbf{Penalty}} \\ \hline
Torque limits & $-5.0$ & DoF position & $-10.0$ \\ 
Dof Velocity & $-5.0$ & Termination & $-200.0$ \\ \hline
\multicolumn{4}{c}{\textbf{Regularization}} \\ \hline
Torque & $-0.000001$ & Action rate & $-0.5$ \\
Feet orientation & $-2.0$ & Feet heading alignment & $-0.02$ \\
Feet air time & $-10.0$ & Slippage & $-1.0$ \\
Hip pos & $-1.0$ & Close feet distance & -0.5 \\ \hline 
\end{tabular}
}
\label{tab:reward}
\vspace{-18pt}
\end{table}

\begin{table}[tbp]
\centering
\vspace{11pt}
\caption{Domain Randomization}
\vspace{-1pt}
\renewcommand{\arraystretch}{1.5}
\resizebox{0.98\linewidth}{!}{%
\begin{tabular}{ c  c  c  c}
\hline
Term & Value & Term & Value \\ \hline
\multicolumn{4}{c}{\textbf{Dynamics}} \\ \hline
Base CoM offset & $\mathcal{U}(-0.1,0.1)$ & Link mass & $\mathcal{U}(0.9,1.1)\times$default \\
Friction & $\mathcal{U}(0.5,1.2)$ & P gain & $\mathcal{U}(0.9, 1.1)\times$default\\
D gain & $\mathcal{U}(0.9,1.1) \times $default & Torque RFI~\cite{campanaro2024learning} & $0.05 \times$torque limit \\ \hline
\multicolumn{4}{c}{\textbf{External Perturbations}} \\ \hline
Push interval & $[5,10]$ & Max push velocity & $1.0$ \\ \hline
\end{tabular}
}
\label{tab:dr}
\vspace{-15pt}
\end{table}

\noindent \textbf{Reference State Initialization and Early Termination.} Proper initialization is crucial for motion imitation. We employ the Reference State Initialization (RSI) framework~\cite{peng2018deepmimic}, where the starting point of the reference motion is randomly sampled and the robot's initial state is derived from the corresponding reference pose. To facilitate efficient training, we implement early termination to avoid collecting invalid data. Unlike previous works that rely on multiple termination conditions (\eg, gravity, height), we simplify the conditions to a single tracking tolerance: the episode terminates if the average link distance between the robot and reference pose exceeds certain threshold. This design avoids direct termination when the gravity or height termination conditions are triggered after RSI (\eg getting up from the ground).

\noindent \textbf{Hard Negative Mining and Motion Filtering.} When training on large datasets, the motion imitation policy may converge to an average point, thereby hindering full coverage of the whole dataset. To address this issue, we employ the strategy of hard negative mining by periodically evaluating our policy over the entire dataset and dynamically adjusting the sampling probability for each motion sample. If the policy fails to track a particular sample, its sampling probability is increased by a predefined factor, whereas successful tracking leads to a corresponding decrease. When the policy's success rate over the entire dataset plateaus and ceases to improve, we apply a filtering mechanism to the original motion dataset. This process identifies samples that persistently fail to be learned, classifying them as implausible instances beyond the capabilities of the current proxy agent. 

\section{BFM Pretraining}

\subsection{BFM under Reinforcement Learning Formulation}

The Behavior Foundation Model is a generative model tasked with learning the underlying distribution of demonstrated behaviors, $P(\tau)$. Under the Markov assumption, the pretraining objective reduces to maximizing the expected log-likelihood over the dataset $\mathcal{D}=\{(s_i^{p,real},a_i)\}_{i=1}^M$ of $M$ state-action pairs by optimizing the model parameters $\theta$:
{\small
\begin{align}
   \max_{\theta} E_{(s_t^{p,real},a_t)\sim \mathcal{D}}[\log \pi_{\theta}(a_t|s_t^{p,real})]
\end{align}}

A monolithic policy $\pi_\theta(a_t|s_t^{p,real})$ is ineffective for control. We introduce the goal state $s_t^{g,real}$ and express the policy as a marginalization over possible goal states in the dataset:
{
\small
\begin{equation}
\begin{split}
    \log \pi_{\theta}(a_t|s_t^{p,real}) 
    = \log &E_{s_t^{g,real}\sim p(s_t^{g,real}|s_t^{p,real})}\\&[\pi_{\theta}(a_t|s_t^{p,real},s_t^{g,real})]
\end{split}
\end{equation}}

By applying the Jensen Inequality, we may obtain a tractable lower bound as a surrogate of the original objective:
{\small
\begin{equation}
\begin{split}
    &\log E_{s_t^{g,real}\sim p(s_t^{g,real}|s_t^{p,real})}[\pi_{\theta}(a_t|s_t^{p,real},s_t^{g,real})] \\
    \ge &E_{s_t^{g,real}\sim p(s_t^{g,real}|s_t^{p,real})}[\log \pi_{\theta}(a_t|s_t^{p,real},s_t^{g,real})]    
\end{split}
\end{equation}
}

The complete pretraining objective for our BFM is then to maximize the lower bound over the entire dataset:
{\small
\begin{equation}
    \begin{split}
        \max_{\theta} E_{(s_t^{p,real},a)\sim \mathcal{D}}&[E_{s_t^{g,real}\sim p(s_t^{g,real}|s_t^{p,real})}\\
        &[\log \pi_{\theta}(a_t|s_t^{p,real},s_t^{g,real})]]
    \end{split}
\end{equation}
}

\subsection{Real-world Proprioception State Design}

The real-world proprioception is defined as $s_t^{p,real}\triangleq [q_{t-25:t}, \dot{q}_{t-25:t}, w_{t-25:t}^{root}, g_{t-25:t},a_{t-25:t-1}]$ which contains the joint position $q_t$, joint velocity $\dot{q_t}$, root angular velocity $w_t^{root}$, projected gravity $g_t$ and the last action $a_{t-1}$. We stack these terms over the last 25 steps to represent proprioception.

\subsection{Control Interface and Mask Strategy}

The distribution $p(s_t^{g,real}|s_t^{p,real})$ highly depends on how you collect and organize the pretraining dataset where goal states $s_t^{g,real}$ are introduced for each state-action pair $(s_t^{p,real},a_t)$. One behavior may be specified by diverse control modes, resulting distinct goal states for current proprioception. For example, as is presented in figure \ref{fig:firstpage}.c, textual instruction, velocity command, whole-body joint positions as well as their arbitrary combinations constitute distinct control modes for specifying the behavior of walking forward. By activating diverse control modes, we actually draw goal state samples from the distribution $p(s_t^{g,real}|s_t^{p,real})$, which allows us to estimate the expectation over goal state distribution.

To simplify matters, we focus on low-level control modes that directly specifies the target state for root, kinematic positions and joint angles and design a control interface compatible with all these control modes, which contains:
\begin{itemize}
    \item Root Control: target root translation, orientation (specified by RPY), linear velocity and angular velocity;
    \item Kinematic Position Control: target rigid-body positions of links rotated to the local frame of reference pose;
    \item Joint Angle Control: target joint angles for each motor;
\end{itemize}
By applying bit-wise binary mask to the control interface, we may activate distinct control modes for low-level control, allowing flexible and versatile control across diverse tasks.

In contrast to prior works~\cite{he2025hover} which adopt a two-stage mask strategy, we directly sample each element of the mask from a Bernoulli distribution $\mathcal{B}(0.5)$, facilitating application of arbitrary control modes. To ensure stable pretraining, we introduce a mask curriculum as a cold-start approach. The sampling probability for each Bernoulli trial gradually decays from an initial value of 1.0 towards 0.5 when the average episode length exceeds a predefined threshold. We adopt a relatively large decay factor in practice, resulting the cold-start phase to span only several hundred episodes.

\subsection{Modeling BFM with Conditional Variational Autoencoder}

We adopt a Conditional Variational Autoencoder (CVAE) to model the log-probability $\log P(a_t|s_t^{p,real},s_t^{g,real})$. The Evidence Lower Bound (ELBO) of CVAE is expressed as:
{\small
\begin{equation}
    \begin{split}
        &E_{q(z|s_t^{p, sim},s_t^{g, sim})}[\log P(a_t|s_t^{p,real}, s_t^{g, real}, z) \\
    &- D_{KL}[q(z|s_t^{p,sim},s_t^{g,sim})||P(z|s_t^{p,real},s_t^{g,real})]]
    \end{split}
\end{equation}
}

We model the prior $\rho$, encoder $\epsilon$ and decoder $D$ as Gaussian distributions and for the decoder, we assume it has a fixed variance. To encourage the latent space to encode more behavioral knowledge, we remove $s_t^{g,real}$ from the input of decoder. Following previous works~\cite{yao2022controlvae, tessler2024maskedmimic}, we design the encoder to be a residual to the prior and include current mask $m_t$ into the encoder input, which can be expressed as:
{\small
\begin{gather}
       P(z|
       s_t^{p,real},s_t^{g,real})= \mathcal{N}(\mu^\rho(s_t^{p,real},s_t^{g,real}),\sigma^\rho (s_t^{p,real},s_t^{g,real})) \\
   q(z|s_t^{p,sim},s_t^{g,sim})=\mathcal{N}(\mu^{\epsilon}(s_t^{p,sim},s_t^{g,sim},m_t) +\mu^\rho(s_t^{p,real},s_t^{g,real}),\notag\\ \sigma^{\epsilon}(s_t^{p,sim},s_t^{g,sim})) \\
    P(a_t|s_t^{p,real},s_t^{g,real},z)=\mathcal{N}(\mu^D(s_t^{p,real},z),\sigma_{{fixed}})
\end{gather}
}

\subsection{Online Distillation}

As we have prepared our behavioral dataset as a proxy agent, we employ the DAgger framework~\cite{ross2011reduction} to optimize the objective of BFM's pretraining. Specifically, for each episode, we roll out the current BFM $\pi_\theta(a_t|s_t^{p,real},s_t^{g,real})$ in simulation to obtain trajectories of $(s_t^{p,real},s_t^{g,real})$. At each timestep, we also compute the corresponding privileged states $(s_t^{p,sim},s_t^{g,sim})$ and query the proxy agent for the reference action $\hat{a_t}$. Parameters of BFM is then updated by:
{
\small
\begin{gather}
L_{DAgger} = ||\hat{a_t} - a_t||_2^2 \notag\\
L_{KL} = D_{KL}(q_{\epsilon}(z_t|s_t^{p,sim},s_t^{g,sim})||P_{\rho}(z|s_t^{p,real},s_t^{g,real})) \notag\\
L = L_{DAgger} + \lambda_{KL}L_{KL}
\end{gather}
}where $\hat{a_t}$ is the reference action from proxy agent, $a_t$ is the action taken by current BFM, $D_{KL}$ is the KL-Divergence operator and $\lambda_{KL}$ maintains balance between the reconstruction quality and the latent space structural regularization. Domain randomization, termination conditions and hard negative mining strategy remain the same as proxy agent training.

\section{BFM Application and Experimental Results}

\begin{table*}[tbp]
\caption{Simulation evaluation of our BFM and baselines on \textbf{VR teleoperation} and \textbf{motion tracking} task. The most significant results are highlighted in bold and wrapped by dark background color and the second significant results are wrapped by light background color. For behavior modulation on motion tracking task, the results with the highest increments relative to our BFM are highlighted in bold and wrapped by dark background color. The results with increments relative to our BFM but not the most are wrapped by light background color. 
}
\label{tab:track_vr}
\centering
\resizebox{\linewidth}{!}{%
\begingroup
\setlength{\tabcolsep}{3pt} 
\renewcommand{\arraystretch}{1} 
\begin{tabular}{lcccccccccccc}
\toprule
\multicolumn{1}{c}{} & \multicolumn{4}{c}{{\textbf{AMASS Train}}} & \multicolumn{4}{c}{{\textbf{AMASS Test}}} & \multicolumn{4}{c}{{\textbf{100Style}}} \\ 
\cmidrule(lr){1-1} \cmidrule(lr){2-5} \cmidrule(lr){6-9} \cmidrule(lr){10-13} 

Method & $E_\text{mpjpe} \downarrow$ & $E_\text{mpkpe} \downarrow$ & $E_\text{lin} \downarrow$ & $\text{E}_{\text{ang}} \downarrow$ & $E_\text{mpjpe} \downarrow$ & $E_\text{mpkpe} \downarrow$ & $E_\text{lin} \downarrow$ & $E_\text{ang}\downarrow$ & $E_\text{mpjpe}\downarrow$ & $E_\text{mpkpe}\downarrow$ & $E_\text{lin}\downarrow$ & $E_\text{ang}\downarrow$ \\ 
\cmidrule(lr){1-1} \cmidrule(lr){2-5} \cmidrule(lr){6-9} \cmidrule(lr){10-13} 

Proxy Agent & 
0.1864 & 49.3057 & 0.1469 & 0.9978 & 
0.2137 & 56.1755 & 0.2631 & 1.3976 & 
0.2460 & 64.1346 & 0.2036 & 1.2336\\
\cmidrule(lr){1-1} \cmidrule(lr){2-5} \cmidrule(lr){6-9} \cmidrule(lr){10-13} 

\multicolumn{13}{l}{\textbf{VR Teleoperation}} \\
\cmidrule(lr){1-1} \cmidrule(lr){2-5} \cmidrule(lr){6-9} \cmidrule(lr){10-13} 

Specialist &
\cellcolor{tablepurpledark}\textbf{0.2113} & \cellcolor{tablepurpledark}\textbf{65.4214} & \cellcolor{tablepurpledark}\textbf{0.2375} & \cellcolor{tablepurpledark}\textbf{1.0925} & 
\cellcolor{tablepurple}0.2555 & \cellcolor{tablepurple}80.5919 & \cellcolor{tablepurple}0.4779 & \cellcolor{tablepurple}1.5036 &
\cellcolor{tablepurpledark}\textbf{0.3062} & \cellcolor{tablepurple}89.9115 & \cellcolor{tablepurpledark}\textbf{0.3189} & \cellcolor{tablepurple}1.1525\\

HOVER &
0.2676 & 91.2667 & 0.5047 & 1.6988 &
0.3055 & 102.8428 & 0.6468 & 1.8716 &
0.3455 & 119.8896 & 0.5351 & 1.6553 \\

BFM (RL from Scratch) &
1.0516 & 399.6902 & 0.4976 & 2.0627 & 
1.1672 & 403.8327 & 0.6528 & 2.3211 &
1.1300 & 429.2893 & 0.4418 & 1.6848 \\

BFM (Ours) &
\cellcolor{tablepurple}0.2447 & \cellcolor{tablepurple}72.3615 & \cellcolor{tablepurple}0.4006 & \cellcolor{tablepurple}1.2177 &
\cellcolor{tablepurpledark}\textbf{0.2235} & \cellcolor{tablepurpledark}\textbf{63.1388} & \cellcolor{tablepurpledark}\textbf{0.3066} & \cellcolor{tablepurpledark}\textbf{1.4632} &
\cellcolor{tablepurple}0.3169 & \cellcolor{tablepurpledark}\textbf{87.0725} & \cellcolor{tablepurple}0.3238 & \cellcolor{tablepurpledark}\textbf{1.1361} \\
\cmidrule(lr){1-1} \cmidrule(lr){2-5} \cmidrule(lr){6-9} \cmidrule(lr){10-13}

\multicolumn{13}{l}{\textbf{Motion Tracking}} \\
\cmidrule(lr){1-1} \cmidrule(lr){2-5} \cmidrule(lr){6-9} \cmidrule(lr){10-13} 

Specialist &
\cellcolor{tablebluedark}\textbf{0.1895} & \cellcolor{tableblue}53.9515 & \cellcolor{tableblue}0.1586 & \cellcolor{tableblue}1.0268 & 
\cellcolor{tableblue}0.2247 & \cellcolor{tableblue}73.6332 & \cellcolor{tablebluedark}\textbf{0.3034} & \cellcolor{tableblue}1.4685 &
\cellcolor{tablebluedark}\textbf{0.2491} & \cellcolor{tableblue}67.7765 & \cellcolor{tableblue}0.2128 & \cellcolor{tablebluedark}\textbf{1.2411} \\

HOVER &
0.2010 & 65.9742 & 0.2189 & 1.1599 &
0.2416 & 87.0678 & 0.3749 & 1.6554 &
\cellcolor{tableblue}0.2562 & 73.9817 & 0.2608 & 1.3369 \\

BFM (RL from Scratch) &
1.0503 & 400.1505 & 0.4973 & 2.0590 & 
1.1689 & 404.7451 & 0.6533 & 2.3532 &
1.1215 & 429.5739 & 0.4422 & 1.6933 \\

BFM (Ours) &
\cellcolor{tableblue}0.1920 & \cellcolor{tablebluedark}\textbf{51.8372} & \cellcolor{tablebluedark}\textbf{0.1542} & \cellcolor{tablebluedark}\textbf{1.0142} &
\cellcolor{tablebluedark}\textbf{0.2226} & \cellcolor{tablebluedark}\textbf{61.1236} & \cellcolor{tableblue}0.3051 & \cellcolor{tablebluedark}\textbf{1.4358} &
0.2637 & \cellcolor{tablebluedark}\textbf{66.4027} & \cellcolor{tablebluedark}\textbf{0.2072} & \cellcolor{tableblue}1.2790 \\
\cmidrule(lr){1-1} \cmidrule(lr){2-5} \cmidrule(lr){6-9} \cmidrule(lr){10-13} 

\multicolumn{13}{l}{\textcolor{red}{\textbf{Behavior Modulation on Motion Tracking}}} \\
\cmidrule(lr){1-1} \cmidrule(lr){2-5} \cmidrule(lr){6-9} \cmidrule(lr){10-13} 

$\lambda=0.5$ &
\cellcolor{tablepeach}0.1893 & \cellcolor{tablepeach}50.4801 & 0.1564 & 1.0419 &
0.2227 & \cellcolor{tablepeach}58.9844 & \cellcolor{tablepeachdark}\textbf{0.2767} & \cellcolor{tablepeachdark}\textbf{1.4099} &
\cellcolor{tablepeach}0.2583 & \cellcolor{tablepeach}63.0582 & 0.2116 & 1.3394 \\

$\lambda=1.0$ &
\cellcolor{tablepeach}0.1875 & \cellcolor{tablepeachdark}\textbf{49.8647} & 0.1609 & 1.0681 &
\cellcolor{tablepeachdark}\textbf{0.2223} & \cellcolor{tablepeachdark}\textbf{58.7251} & \cellcolor{tablepeach}0.2870 & 1.4899 & 
\cellcolor{tablepeachdark}\textbf{0.2562} & \cellcolor{tablepeachdark}\textbf{62.5168} & 0.2224 & 1.3919 \\

$\lambda=1.5$ &
\cellcolor{tablepeachdark}\textbf{0.1869} & \cellcolor{tablepeach}50.1451 & 0.1675 & 1.0866 &
\cellcolor{tablepeach}0.2224 & \cellcolor{tablepeach}60.4565 & \cellcolor{tablepeach}0.2990 & 1.4964 &
\cellcolor{tablepeach}0.2567 & \cellcolor{tablepeach}64.0520 & 0.2370 & 1.4618 \\

$\lambda=2.0$ &
0.2625 & 76.2392 & 0.2615 & 1.5176 &
0.2254 & 67.6583 & 0.3158 & 1.5438 &
\cellcolor{tablepeach}0.2625 & 76.2392 & 0.2615 & 1.5176 \\
\cmidrule(lr){1-1} \cmidrule(lr){2-5} \cmidrule(lr){6-9} \cmidrule(lr){10-13}

\bottomrule 
\end{tabular}
\endgroup
}
\vspace{-16pt}
\end{table*}

\begin{table}[tbp]
\caption{Simulation Evaluation of BFM and baselines on \textbf{locomotion} task across three datasets. The most significant results are highlighted in bold and wrapped by dark background color and the second significant results are wrapped by light background color.
}
\label{tab:loco}
\centering
\resizebox{0.98\linewidth}{!}{%
\begingroup
\setlength{\tabcolsep}{3pt} 
\renewcommand{\arraystretch}{1} 
\begin{tabular}{lcccccc}
\toprule

\multicolumn{1}{c}{} & \multicolumn{2}{c}{{\textbf{AMASS Train}}} & \multicolumn{2}{c}{{\textbf{AMASS Test}}} & \multicolumn{2}{c}{{\textbf{100Style}}} \\ 
\cmidrule(lr){1-1} \cmidrule(lr){2-3} \cmidrule(lr){4-5} \cmidrule(lr){6-7} 

Experiment & $E_\text{lin,xy}\downarrow$ & $E_\text{ang,z}\downarrow$ & $E_\text{lin,xy}\downarrow$ & $E_\text{ang,z}\downarrow$ & $E_\text{lin,xy}\downarrow$ & $E_\text{ang,z}\downarrow$\\ 
\cmidrule(lr){1-1} \cmidrule(lr){2-3} \cmidrule(lr){4-5} \cmidrule(lr){6-7} 

Specialist & \cellcolor{tablebluedark}\textbf{0.1201} & \cellcolor{tablebluedark}\textbf{0.4801} & \cellcolor{tableblue}0.2168 & \cellcolor{tableblue}0.6751 & \cellcolor{tablebluedark}\textbf{0.1496} & \cellcolor{tableblue}0.5108 \\
HOVER & 0.1494 & 0.5518 & 0.2663 & 0.7624 & 0.1696 & 0.5707 \\
BFM (RL from Scratch) & 0.4314 & 1.2925 & 0.5513 & 1.4982 & 0.4015 & 1.0606 \\
BFM (Ours) & \cellcolor{tableblue}0.1292 & \cellcolor{tableblue}0.4974 & \cellcolor{tablebluedark}\textbf{0.2116} & \cellcolor{tablebluedark}\textbf{0.6744} & \cellcolor{tableblue}0.1603 & \cellcolor{tablebluedark}\textbf{0.4973} \\
\cmidrule(lr){1-1} \cmidrule(lr){2-3} \cmidrule(lr){4-5} \cmidrule(lr){6-7} 

\bottomrule
\end{tabular}
\endgroup
}
\vspace{-2em}
\end{table}

\subsection{Experiment Setup}

The training of our proxy agent and BFM is conducted in IsaacGym~\cite{makoviychuk2021isaac}, with 8192 parallel environments. To ensure both the efficiency and persuasiveness of our evaluation, we report metrics calculated based on IsaacGym and demonstrate both the sim-to-sim results in Mujoco~\cite{todorov2012mujoco} and sim-to-real results in real world. We adopt the Unitree G1 humanoid robot~\cite{unitreeg1} as an agile and powerful platform for real-world deployment, which stands 1.3 meters tall and has 29 degrees of freedom. To simplify the difficulty of control, we freeze the wrists of both hands, resulting in 23 degrees of freedom.

\subsection{Steering BFM with the Control Interface}

As our BFM enables humanoid control via diverse control modes, we first demonstrate its application of direct steering for multiple WBC tasks.  We select three prevailing WBC tasks: whole-body motion tracking, VR teleoperation and locomotion to demonstrate the effectiveness of our BFM. For each task, we activate the corresponding control mode by manually crafting and applying mask to the control interface.

\noindent\textbf{Baselines.} To prove that our BFM has encoded extensive behavioral knowledge that can be directly steered by diverse WBC tasks, we select HOVER\cite{he2025hover} as a general baseline for all the three tasks. For each task, we also select a specialist to show that our BFM is \textit{as good as, if not better than}, the specialists for they often indicate overfitting to a specific control mode. We also train a BFM with the same architecture and hyper-parameters from scratch with reinforcement learning to ablate our options on the learning paradigm and the results of proxy agent are also included in the table for motion tracking and teleoperation. We select specialists with the same online distillation process and follows their implementation to align the training details which might vary from embodiment to dataset. For whole-body motion tracking, we follow the implementation of GMT~\cite{chen2025gmt} and for VR teleoperation, we refer to the implementation of OmniH2O~\cite{he2024omnih2o}. We use our own implementation of specialists for the Locomotion task. 

\noindent\textbf{Metrics.} For whole-body motion tracking and VR teleoperation, their goal is to track the control signals while demonstrating whole-body coordination. Therefore we adopt the same metric set for these two tasks comprised of the mean per-keypoint error (MPKPE) $E_{mpkpe}(mm)$, mean per-joint error (MPJPE) $E_{mpjpe}(rad)$, root linear velocity tracking error $E_{lin}(m/s)$ and angular velocity tracking error $E_{ang}(rad/s)$. For locomotion, its goal reduces to following a velocity command specified by root linear velocity on xy-plane and angular velocity along the z-axis. As a consequence, we adopt a metric set which contains the root linear velocity error on xy-plane $E_{lin,xy}$ and the root angular velocity error along z-axis $E_{ang,z}$. All the metrics are evaluated on three datasets, the training set of AMASS, the test set of AMASS and the 100STYLE dataset~\cite{mason2022local}. 

\noindent\textbf{Experimental Results.} As is demonstrated in Table \ref{tab:track_vr} and \ref{tab:loco}, our BFM consistently outperforms HOVER on almost all metrics across all the tasks and datasets. Also, our BFM is as good as, if no better than the specialists. We attribute the conditions where specialists may outperform our BFM to two reasons: 1) the specialists focus on a specific control mode, naturally allowing better learning of an unchanged setting. 2) the specialists may overfit to the training set under specific control mode. Besides, we observe that our BFM consistently outperforms the RL policy trained from scratch, which confirms that our training paradigm is effective for BFM pretraining. The overall results highlight our BFM's versatility and generalization ability across multiple tasks.

\subsection{Behavior Composition and Modulation with BFM}

\begin{figure*}[tbp]
    \centering
    \includegraphics[width=0.91\textwidth]{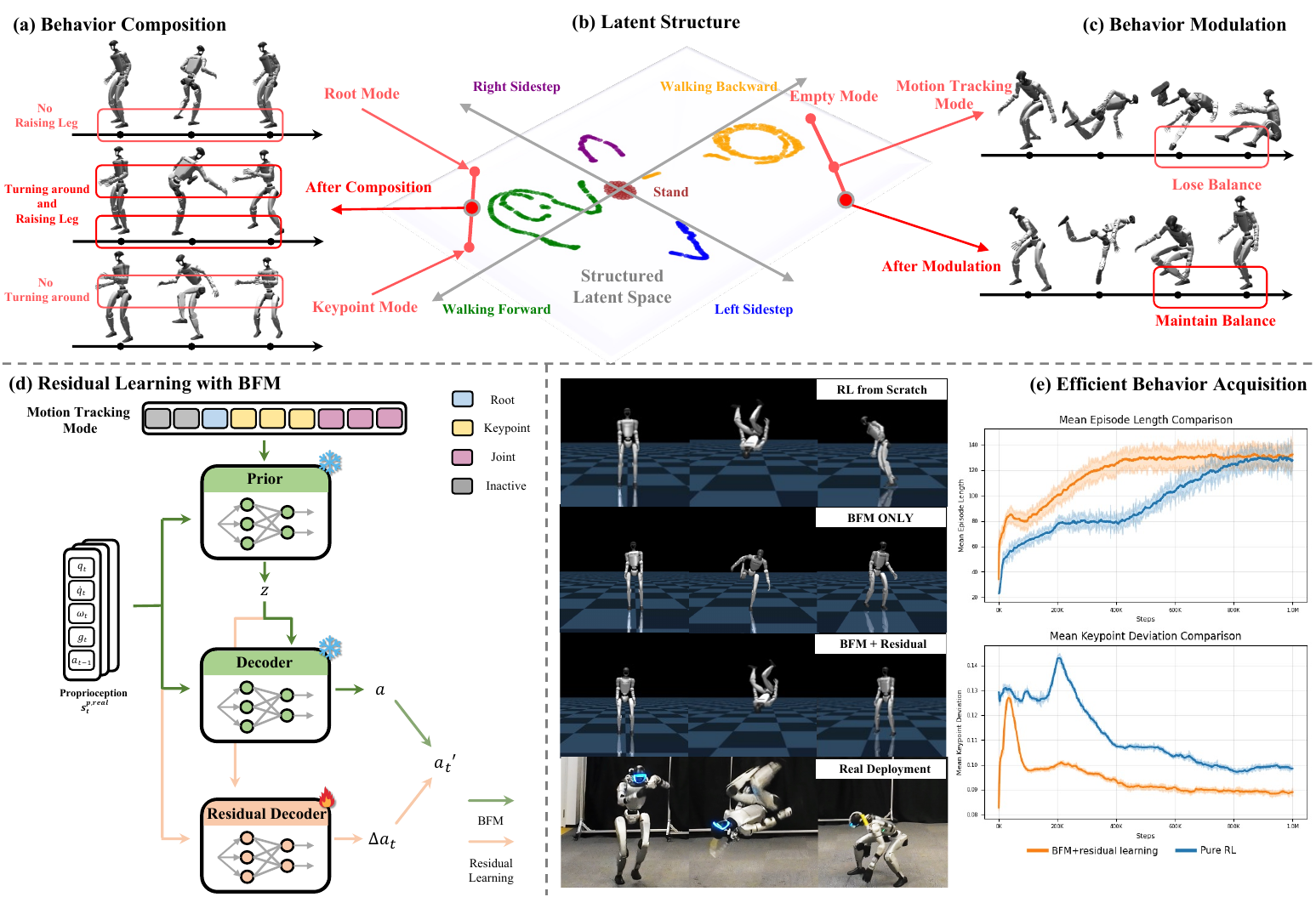}
    \caption{(a) BFM allows behavior composition for novel behaviors through linear interpolation of latent variables from two distinct control modes. (b) T-SNE results (brown for standing, green for walking forward, yellow for walking backward, blue for left sidestep and purple for right sidestep) demonstrates our BFM provides a structured latent space with clear directionality and symmetry. (c) BFM allows behavior modulation to better align with desired control modes through linear extrapolation in the latent space. (d) We adopt residual learning on our pretrained BFM for efficient acquisition of novel behaviors. (e) We select Side Salto as a challenging case and compare our method with training RL policy from scratch.}
    \label{fig:exp_main}
    \vspace{-2em}
\end{figure*}

One key advantage of using a CVAE to model the BFM is that it provides a structured latent space that encodes a broad spectrum of behavioral knowledge. To fully unleash the potential of our BFM, we first perform latent analysis to clarify how the latent space has been structured. Then based on the analysis results, we further perform experiments to demonstrate some unique and special properties of our BFM.

\noindent\textbf{Latent Structure.} We first choose five motions including standing still, walking forward, walking backward, left sidestep and right sidestep. Then we adopt the control mode of motion tracking to collect latent sequences and apply the t-SNE algorithm~\cite{maaten2008visualizing} to project the high-dimensional latent variables into a 2D plane for visualization. 
As shown in figure \ref{fig:exp_main}.b, we observe that 1) the projected latent variables demonstrate clear directionality and symmetry. 2) While humanoids are all initialized as the same standing pose, the latent variables are pre-clustered instead of transiting from center to diverse directions, indicating our model may have learned strong prior over latent space. Overall, this analysis confirms that the CVAE learns a meaningful and structured latent manifold, which can be leveraged for diverse purposes.

\noindent\textbf{Behavior Composition.} We explore the possibility of interpolation in the latent space for producing novel behaviors. We select the Roundhouse Kick as a difficult motion to perform and activates the control mode of root and keypoint separately. We observe that root-only control leads to a turning movement without raising leg while keypoint-only control results in raising leg without turning around. However, when we linearly interpolate the latent variables from these two control modes with a coefficient of 0.5, the humanoid could complete the Roundhouse Kick motion, as is demonstrated in figure \ref{fig:exp_main}.a. By gradually increasing the coefficient from 0 to 1, the humanoid exhibits a clear transition among root-only control, full completion and keypoint-only control. The overall results highlight the structure our BFM has acquired which allows flexible compositions of diverse behaviors.

\noindent\textbf{Behavior Modulation.} We further explore the possibility of extrapolation in the latent space for better alignment with desired mode. We select the Butterfly Kick as a challenging motion the model fails to directly perform under the motion tracking mode, where it will lose its balance when landing on the ground. We propose to obtain latent variables in a similar way to Classifier-free Guidance~\cite{ho2022classifier} in Diffusion models:
{\small
\begin{align}
    z = (1+\lambda)\mu^{\rho}(s_t^{p,real},s_t^{g,real}) - \lambda\mu^{\rho}(s_{t}^{p,real},\emptyset), \lambda>0
\end{align}}
We find that by setting $\lambda$ as 0.5, the humanoid can now maintain its balance when landing, as is demonstrated in figure \ref{fig:exp_main}.c. To further clarify its effect, we apply this finding to the motion tracking task. As is presented in table \ref{tab:track_vr}, when applying a medium coefficient, the tracking results all achieve improvements to some extent. While a relatively large coefficient may lead to degradation of performance, indicating an excessive modulation towards the control mode.

\vspace{-0.5em}
\subsection{Efficient Behavior Acquisition with BFM}
By pretraining over large-scale behavioral dataset, our BFM is equipped with sufficient behavioral knowledge for efficient acquisition of novel behaviors. By focusing on novel behaviors specified by reference motions, we still adopt motion imitation as an effective paradigm for behavior learning. Specifically, we freeze all the parameters of BFM and activate the control mode of motion tracking. Upon our pretrained model, we learn a residual decoder $\pi(\Delta a_t|s_t^{p,real},z)$ and the final action becomes $a_t'=a_t+\Delta a_t$, as is presented in figure \ref{fig:exp_main}.d. The training of residual model follows the same configuration as the proxy agent, except we activate termination curriculum~\cite{he2025asap} and adjust the threshold based on the sequence length of each motion. To demonstrate the effectiveness of residual learning, we select side salto as a challenging motion that the BFM can not directly handle and present the visualization results and curves for mean keypoint deviations and mean episode lengths. As is shown in figure \ref{fig:exp_main}.e, by comparing our methods of residual learning on BFM with methods of learning RL policy from scratch, the existence of BFM avoids inefficient exploration at early stage of training and achieves more accurate tracking results based on the behavioral knowledge our BFM has learned. 

\section{CONCLUSION}
In this work, we introduce Behavior Foundation Model for humanoid robots, a generative model for behaviors pretrained on large-scale behavioral dataset to encode extensive, reusable behavioral knowledge. Based on mathematical analysis under RL formulation, we implement our BFM through the training of a proxy agent and online masked distillation by a CVAE. Comprehensive evaluations consolidate that our BFM achieves strong capabilities of cross-task generalization, behavior composition, behavior modulation and efficient acquisition of novel behaviors. Future work may further extend the current simplified control interface to support a broader range of control modes.

\printbibliography

@article{he2024omnih2o,
  title={Omnih2o: Universal and dexterous human-to-humanoid whole-body teleoperation and learning},
  author={He, Tairan and Luo, Zhengyi and He, Xialin and Xiao, Wenli and Zhang, Chong and Zhang, Weinan and Kitani, Kris and Liu, Changliu and Shi, Guanya},
  journal={arXiv preprint arXiv:2406.08858},
  year={2024}
}

@article{ze2025twist,
  title={TWIST: Teleoperated Whole-Body Imitation System},
  author={Ze, Yanjie and Chen, Zixuan and Ara{\~A}{\v{s}}jo, Jo{\~A}{\c{G}}o Pedro and Cao, Zi-ang and Peng, Xue Bin and Wu, Jiajun and Liu, C Karen},
  journal={arXiv preprint arXiv:2505.02833},
  year={2025}
}

@article{chen2025gmt,
  title={GMT: General Motion Tracking for Humanoid Whole-Body Control},
  author={Chen, Zixuan and Ji, Mazeyu and Cheng, Xuxin and Peng, Xuanbin and Peng, Xue Bin and Wang, Xiaolong},
  journal={arXiv preprint arXiv:2506.14770},
  year={2025}
}

@article{radford2019language,
  title={Language models are unsupervised multitask learners},
  author={Radford, Alec and Wu, Jeffrey and Child, Rewon and Luan, David and Amodei, Dario and Sutskever, Ilya and others},
  journal={OpenAI blog},
  volume={1},
  number={8},
  pages={9},
  year={2019}
}

@article{tessler2024maskedmimic,
  title={Maskedmimic: Unified physics-based character control through masked motion inpainting},
  author={Tessler, Chen and Guo, Yunrong and Nabati, Ofir and Chechik, Gal and Peng, Xue Bin},
  journal={ACM Transactions on Graphics (TOG)},
  volume={43},
  number={6},
  pages={1--21},
  year={2024},
  publisher={ACM New York, NY, USA}
}

@inproceedings{he2025hover,
  title={Hover: Versatile neural whole-body controller for humanoid robots},
  author={He, Tairan and Xiao, Wenli and Lin, Toru and Luo, Zhengyi and Xu, Zhenjia and Jiang, Zhenyu and Kautz, Jan and Liu, Changliu and Shi, Guanya and Wang, Xiaolong and others},
  booktitle={2025 IEEE International Conference on Robotics and Automation (ICRA)},
  pages={9989--9996},
  year={2025},
  organization={IEEE}
}

@inproceedings{higgins2017beta,
  title={beta-vae: Learning basic visual concepts with a constrained variational framework},
  author={Higgins, Irina and Matthey, Loic and Pal, Arka and Burgess, Christopher and Glorot, Xavier and Botvinick, Matthew and Mohamed, Shakir and Lerchner, Alexander},
  booktitle={International conference on learning representations},
  year={2017}
}

@inproceedings{he2016deep,
  title={Deep residual learning for image recognition},
  author={He, Kaiming and Zhang, Xiangyu and Ren, Shaoqing and Sun, Jian},
  booktitle={Proceedings of the IEEE conference on computer vision and pattern recognition},
  pages={770--778},
  year={2016}
}

@article{schulman2017proximal,
  title={Proximal policy optimization algorithms},
  author={Schulman, John and Wolski, Filip and Dhariwal, Prafulla and Radford, Alec and Klimov, Oleg},
  journal={arXiv preprint arXiv:1707.06347},
  year={2017}
}

@inproceedings{mahmood2019amass,
  title={AMASS: Archive of motion capture as surface shapes},
  author={Mahmood, Naureen and Ghorbani, Nima and Troje, Nikolaus F and Pons-Moll, Gerard and Black, Michael J},
  booktitle={Proceedings of the IEEE/CVF international conference on computer vision},
  pages={5442--5451},
  year={2019}
}

@incollection{loper2023smpl,
  title={SMPL: A skinned multi-person linear model},
  author={Loper, Matthew and Mahmood, Naureen and Romero, Javier and Pons-Moll, Gerard and Black, Michael J},
  booktitle={Seminal Graphics Papers: Pushing the Boundaries, Volume 2},
  pages={851--866},
  year={2023}
}

@inproceedings{he2024learning,
  title={Learning human-to-humanoid real-time whole-body teleoperation},
  author={He, Tairan and Luo, Zhengyi and Xiao, Wenli and Zhang, Chong and Kitani, Kris and Liu, Changliu and Shi, Guanya},
  booktitle={2024 IEEE/RSJ International Conference on Intelligent Robots and Systems (IROS)},
  pages={8944--8951},
  year={2024},
  organization={IEEE}
}

@inproceedings{campanaro2024learning,
  title={Learning and deploying robust locomotion policies with minimal dynamics randomization},
  author={Campanaro, Luigi and Gangapurwala, Siddhant and Merkt, Wolfgang and Havoutis, Ioannis},
  booktitle={6th Annual Learning for Dynamics \& Control Conference},
  pages={578--590},
  year={2024},
  organization={PMLR}
}

@article{peng2018deepmimic,
  title={Deepmimic: Example-guided deep reinforcement learning of physics-based character skills},
  author={Peng, Xue Bin and Abbeel, Pieter and Levine, Sergey and Van de Panne, Michiel},
  journal={ACM Transactions On Graphics (TOG)},
  volume={37},
  number={4},
  pages={1--14},
  year={2018},
  publisher={ACM New York, NY, USA}
}

@article{yao2022controlvae,
  title={Controlvae: Model-based learning of generative controllers for physics-based characters},
  author={Yao, Heyuan and Song, Zhenhua and Chen, Baoquan and Liu, Libin},
  journal={ACM Transactions on Graphics (TOG)},
  volume={41},
  number={6},
  pages={1--16},
  year={2022},
  publisher={ACM New York, NY, USA}
}

@inproceedings{ross2011reduction,
  title={A reduction of imitation learning and structured prediction to no-regret online learning},
  author={Ross, St{\'e}phane and Gordon, Geoffrey and Bagnell, Drew},
  booktitle={Proceedings of the fourteenth international conference on artificial intelligence and statistics},
  pages={627--635},
  year={2011},
  organization={JMLR Workshop and Conference Proceedings}
}

@article{makoviychuk2021isaac,
  title={Isaac gym: High performance gpu-based physics simulation for robot learning},
  author={Makoviychuk, Viktor and Wawrzyniak, Lukasz and Guo, Yunrong and Lu, Michelle and Storey, Kier and Macklin, Miles and Hoeller, David and Rudin, Nikita and Allshire, Arthur and Handa, Ankur and others},
  journal={arXiv preprint arXiv:2108.10470},
  year={2021}
}

@inproceedings{todorov2012mujoco,
  title={Mujoco: A physics engine for model-based control},
  author={Todorov, Emanuel and Erez, Tom and Tassa, Yuval},
  booktitle={2012 IEEE/RSJ international conference on intelligent robots and systems},
  pages={5026--5033},
  year={2012},
  organization={IEEE}
}

@misc{unitreeg1,
  title = {Unitree g1 humanoid agent ai avatar},
  author = {Unitree},
  url = "https://www.unitree.com/g1",
  year = 2024,
}

@article{mason2022local,
author = {Mason, Ian and Starke, Sebastian and Komura, Taku},
title = {Real-Time Style Modelling of Human Locomotion via Feature-Wise Transformations and Local Motion Phases},
year = {2022},
publisher = {Association for Computing Machinery},
address = {New York, NY, USA},
volume = {5},
number = {1},
doi = {10.1145/3522618},
journal = {Proceedings of the ACM on Computer Graphics and Interactive Techniques},
month = {may},
articleno = {6}
}

@article{maaten2008visualizing,
  title={Visualizing data using t-SNE},
  author={Maaten, Laurens van der and Hinton, Geoffrey},
  journal={Journal of machine learning research},
  volume={9},
  number={Nov},
  pages={2579--2605},
  year={2008}
}

@article{ho2022classifier,
  title={Classifier-free diffusion guidance},
  author={Ho, Jonathan and Salimans, Tim},
  journal={arXiv preprint arXiv:2207.12598},
  year={2022}
}

@article{jiang2024harmon,
  title={Harmon: Whole-body motion generation of humanoid robots from language descriptions},
  author={Jiang, Zhenyu and Xie, Yuqi and Li, Jinhan and Yuan, Ye and Zhu, Yifeng and Zhu, Yuke},
  journal={arXiv preprint arXiv:2410.12773},
  year={2024}
}

@article{shao2025langwbc,
  title={LangWBC: Language-directed Humanoid Whole-Body Control via End-to-end Learning},
  author={Shao, Yiyang and Huang, Xiaoyu and Zhang, Bike and Liao, Qiayuan and Gao, Yuman and Chi, Yufeng and Li, Zhongyu and Shao, Sophia and Sreenath, Koushil},
  journal={arXiv preprint arXiv:2504.21738},
  year={2025}
}

@article{xue2025unified,
  title={A Unified and General Humanoid Whole-Body Controller for Versatile Locomotion},
  author={Xue, Yufei and Dong, Wentao and Liuˆ, Minghuan and Zhang, Weinan and Pang, Jiangmiao},
  journal={arXiv preprint arXiv:2502.03206},
  year={2025}
}

@article{tirinzoni2025zero,
  title={Zero-shot whole-body humanoid control via behavioral foundation models},
  author={Tirinzoni, Andrea and Touati, Ahmed and Farebrother, Jesse and Guzek, Mateusz and Kanervisto, Anssi and Xu, Yingchen and Lazaric, Alessandro and Pirotta, Matteo},
  journal={arXiv preprint arXiv:2504.11054},
  year={2025}
}

@inproceedings{lu2025mobile,
  title={Mobile-television: Predictive motion priors for humanoid whole-body control},
  author={Lu, Chenhao and Cheng, Xuxin and Li, Jialong and Yang, Shiqi and Ji, Mazeyu and Yuan, Chengjing and Yang, Ge and Yi, Sha and Wang, Xiaolong},
  booktitle={2025 IEEE International Conference on Robotics and Automation (ICRA)},
  pages={5364--5371},
  year={2025},
  organization={IEEE}
}

@article{ben2025homie,
  title={Homie: Humanoid loco-manipulation with isomorphic exoskeleton cockpit},
  author={Ben, Qingwei and Jia, Feiyu and Zeng, Jia and Dong, Junting and Lin, Dahua and Pang, Jiangmiao},
  journal={arXiv preprint arXiv:2502.13013},
  year={2025}
}

@article{he2025asap,
  title={Asap: Aligning simulation and real-world physics for learning agile humanoid whole-body skills},
  author={He, Tairan and Gao, Jiawei and Xiao, Wenli and Zhang, Yuanhang and Wang, Zi and Wang, Jiashun and Luo, Zhengyi and He, Guanqi and Sobanbab, Nikhil and Pan, Chaoyi and others},
  journal={arXiv preprint arXiv:2502.01143},
  year={2025}
}

\end{document}